\documentclass[10pt, a4paper]{article}

\usepackage[final]{lrec-coling2024} 

\usepackage{arydshln}

\usepackage{microtype}
\usepackage{amsmath}
\usepackage{algorithm}
\usepackage{algpseudocode}
\usepackage{enumitem}

\title{A Knowledge Plug-and-Play Test Bed for Open-domain Dialogue Generation}

\name{Xiangci Li\textsuperscript{\rm 1}*\thanks{*Work performed while the author was interning at Tencent AI Lab.}  Linfeng Song\textsuperscript{\rm 2}  Lifeng Jin\textsuperscript{\rm 2}  Haitao Mi\textsuperscript{\rm 2}  Jessica Ouyang\textsuperscript{\rm 1}  Dong Yu\textsuperscript{\rm 2}} 

\address{\textsuperscript{\rm 1}The University of Texas at Dallas, Richardson, TX, USA\\ 
\textsuperscript{\rm 2}Tencent AI Lab, Bellevue, WA, USA  \\
         lixiangci8@gmail.com, Jessica.Ouyang@utdallas.edu\\
         \{lfsong, lifengjin, haitaomi, dyu\}@tencent.com\\}

\abstract{Knowledge-based, open-domain dialogue generation aims to build chit-chat systems that talk to humans using mined support knowledge. Many types and sources of knowledge have previously been shown to be useful as support knowledge. Even in the era of large language models, response generation grounded in knowledge retrieved from additional up-to-date sources remains a practically important approach. While prior work using single-source knowledge has shown a clear positive correlation between the performances of knowledge selection and response generation, there are no existing multi-source datasets for evaluating support knowledge retrieval. Further, prior work has assumed that the knowledge sources available at test time are the same as during training. This unrealistic assumption unnecessarily handicaps models, as new knowledge sources can become available after a model is trained. In this paper, we present a high-quality benchmark named \textit{multi-source Wizard of Wikipedia} (Ms.WoW) for evaluating multi-source dialogue knowledge selection and response generation. Unlike existing datasets, it contains clean support knowledge, grounded at the utterance level and partitioned into multiple knowledge sources. We further propose a new challenge, \textit{dialogue knowledge plug-and-play}, which aims to test an already trained dialogue model on using new support knowledge from previously unseen sources in a zero-shot fashion.}

\begin{document}

\maketitleabstract

\section{Introduction}

Knowledge-based open-domain dialogue generation aims to build chit-chat systems that talk to humans on various domains with mined support knowledge.
Many types of knowledge have been shown to be useful as support knowledge, such as encyclopedias \cite{dinan2019wizard}, knowledge graphs \cite{wu-etal-2019-proactive,zhou-etal-2020-kdconv,liu-etal-2021-durecdial}, personas \cite{zhang-etal-2018-personalizing}, and commonsense knowledge \cite{zhou2018commonsense, zhang-etal-2020-grounded,wu2021more, varshney-etal-2022-commonsense}.  

Further, \citet{shuster2022blenderbot} have demonstrated that multiple knowledge sources are helpful on top of large-scale, pre-trained language models. Even in the era of large language models \citep[LLMs;][]{brown2020language, openai2023gpt4, touvron2023llama}, which use internally learned knowledge from their pretraining corpora for zero- or few-shot predictions, knowledge can become outdated. Thus, response generation grounded in knowledge retrieved from additional up-to-date sources is still a practically important approach.

Previous studies using single-source knowledge for response generation have shown a clear positive correlation between the performances of knowledge selection and response generation \cite{dinan2019wizard,li-etal-2022-enhancing-knowledge}. In this work, we aim to extend these results to multi-source knowledge and further propose a new challenge task, \textit{dialogue knowledge plug-and-play}. We address two major challenges facing multi-source knowledge-based dialogue generation. First, there are no existing multi-source datasets for evaluating support knowledge retrieval. Prior work used non-knowledge-based dialogue datasets with silver support knowledge labeled using unsupervised approaches \cite{liu2019knowledge, wu2021more}; they could only measure the final response generation performance, without being able to evaluate the support knowledge selection module. 
As a result, results achieved on these datasets lack interpretability: the relationship between the quality of knowledge selection and response generation is unclear, adding an extra layer of difficulty in improving models' performance. 

Second, prior work has assumed that the knowledge sources available at test time are the same as during training. We argue that this is an unrealistic assumption that unnecessarily handicaps models. New knowledge sources can become available after a model is trained: new knowledge graphs are published, or new types of knowledge are shown to be useful for dialogue generation. Information present only in a new knowledge source (for example, about a recent newsworthy event) may be crucial in a conversation with a real user. To make use of such information, it is necessary to ensure that trained dialogue generation models are robust to the addition of new knowledge sources at inference time: the new source should improve, or at the very least not harm, a model's performance.

To overcome these challenge, we present a high-quality benchmark named \emph{multi-source Wizard of Wikipedia (Ms.WoW)}\footnote{\url{https://github.com/jacklxc/Ms.WoW}} for evaluating multi-source dialogue knowledge selection and response generation. Unlike existing datasets, it contains clean, gold support knowledge, grounded at the utterance level and partitioned into multiple knowledge ``sources." We build Ms.WoW on top of the Wizard of Wikipedia (WoW, \citealt{dinan2019wizard}), which annotates utterance-level, grounded support knowledge sentences. We partition the knowledge in WoW into different ``sources," including OPIEC \cite{gashteovski2019opiec}, semantic frames, and Wikidata, to simulate multiple knowledge sources containing complementary information.

Using the Ms.WoW dataset, we introduce the \emph{dialogue knowledge plug-and-play} challenge task, which aims to test an already trained dialogue model on using new support knowledge from previously unseen sources in a zero-shot fashion. The plug-and-play task extends WoW to the real-world scenario where the knowledge sources available at inference time are different from those available during training. Thus, Ms.WoW is a test bed for both evaluating the effect of multi-source knowledge selection on dialogue response generation, as well as simulating the challenging zero-shot knowledge source adaptation scenario.

\begin{table*}[t]
\begin{center}
\setlength{\tabcolsep}{2.5pt} 
\renewcommand{\arraystretch}{1.0} 
\small
    \begin{tabular}{  p{0.83\linewidth} | l | l  }
    \hline
    \multicolumn{3}{l}{\textbf{Response}} \\ 
    \multicolumn{3}{l}{It was formed in 1965. The Pepsi-Cola company and Frito-Lay merged to form one big company.} \\ 
    \hline \hline
    \multicolumn{3}{l}{\textbf{Gold WoW sentence}} \\ 
    \multicolumn{3}{p{\dimexpr\linewidth-2\tabcolsep\relax}}{PepsiCo was formed in 1965 with the merger of the Pepsi-Cola Company and Frito-Lay, Inc. PepsiCo has since expanded from its namesake product Pepsi to a broader range of food and beverage brands, the largest of which included an acquisition of Tropicana Products in 1998 and the Quaker Oats Company in 2001, which added the Gatorade brand to its portfolio.} \\
    \hline \hline
    \textbf{Ms.WoW Knowledge Tuples} & \textbf{Source} & \textbf{Gold} \\
    \hline
    (`', `formed', `PepsiCo', `in 1965', `') & Sem. frm. & Yes \\
    (`its', `', `has', `namesake product Pepsi', `', `') & OPIEC & No \\
    (`largest of which', `', `have included acquisition of Tropicana Products in 1998', `beverage brands', `', `') & OPIEC & No \\
    (`largest of which', `', `have included Quaker Oats Company in 2001', `food brands', `', `') & OPIEC & No \\
    (`Quaker Oats Company in 2001', `', `added Gatorade brand to', `portfolio', `', `') & OPIEC & Yes \\ 
    (`Frito-Lay', `parent organization', `PepsiCo') & Wikidata & Yes \\
    (`Pepsi', `instance of', `cola') & Wikidata & Yes \\
    \hline
    \end{tabular}
    \vspace{-0.5em}
    \caption{An example decomposition of a gold knowledge sentence from WoW into tuples from multiple sources in Ms.WoW. Tuples not used in the response are not labeled as gold tuples.} \label{tab:example}
    \vspace{-1.5em}
\end{center}
\end{table*}

\section{Background \& Related Work} \label{sec:background}

\subsection{Open-Domain Dialogue Generation}
To the best of our knowledge, no existing open-domain dialogue dataset is well-suited for the study of \textit{dialogue knowledge plug-and-play}. 

First, most knowledge-grounded, open domain dialogue corpora only provide support knowledge from a single source \cite{wu-etal-2019-proactive,zhou-etal-2020-kdconv,liu-etal-2021-durecdial,wang2021naturalconv, komeili2021internet}. For example, the WoW dataset \cite{dinan2019wizard}, which we use as the base for our Ms.WoW, uses a knowledgeable ``wizard" speaker who records the Wikipedia support articles for each utterance during a conversation. However, all of that support knowledge is from a single source: Wikipedia plain text. In this work, we partition the knowledge in WoW to simulate the availability of several ``sources" containing complementary knowledge.

Second, existing multi-source, knowledge-based dialogue generation works collect support knowledge after the conversation occurs, so there is no gold support knowledge grounded for each utterance. \citet{liu2019knowledge} map an unreleased knowledge graph to two existing datasets \cite{moghe-etal-2018-towards,dinan2019wizard} to create an augmented multi-source dataset. \citet{wu2021more} build a single-turn dialogue dataset upon three Weibo corpora \cite{shang-etal-2015-neural,ke-etal-2018-generating,cai-etal-2019-retrieval} by extracting post-reply pairs and further augment the utterance pairs with ConceptNet \cite{speer2017conceptnet}, as well as article text and infobox tables from Chinese Wikipedia. These works are not able to report explicit knowledge selection performance due to the lack of ground-truth support knowledge. Further, their methods produce noisy support knowledge, as erroneous examples are frequently seen in their automatic knowledge-matching process.

Holl-E \cite{moghe-etal-2018-towards} is the closest multi-source dialogue work to ours; however, they focus only on the movie domain, retrieving plot, review, fact, and comment information for each utterance. Moreover, they do not perform knowledge selection, so the correlation between knowledge selection and dialogue generation performance is unknown.

In contrast, the single-source WoW dataset contains gold support knowledge annotated at the utterance level, and because state-of-the-art Transformer-based models are unable to take all candidate knowledge (i.e. all relevant Wikipedia articles) as input due to their length limits, \citet{dinan2019wizard} perform knowledge selection to filter out unneeded candidates and feed only the selected support knowledge to the dialogue response generator. Thus, dialogue systems evaluated on WoW are able to report knowledge selection performance and observe the positive correlation between the performances of knowledge selection and response generation. For these reasons, we build our Ms.WoW on top of WoW.

\subsection{Plug-and-play}
The concept of plug-and-play has been introduced in the context of studying language models' ability to adapt to new knowledge. \citet{DBLP:conf/iclr/DathathriMLHFMY20} proposed a Plug and Play Language Model (PPLM) for controllable language generation. \citet{xu-etal-2021-k-plug} proposed K-PLUG, a knowledge-injected, pre-trained language model for e-commerce that handles information such as item category and attributes in key-value pairs, as well as item summaries and descriptions in plain text. In this work, we use our new Ms.WoW dataset and the task of multi-source dialogue knowledge selection and response generation to study the problem of \textit{dialogue knowledge plug-and-play}.

\section{Multi-Source Wizard of Wikipedia Dataset}

\subsection{Knowledge Tuple Retrieval or Extraction}
We create Ms.WoW by replacing the Wikipedia knowledge sentences in WoW \cite{dinan2019wizard} with tuples retrieved or extracted from three sources with different emphases. The three sources, described in detail below, provide \textit{disjoint partitions} that cover the semantics of the original WoW sentences\footnote{All filtering thresholds in this section are empirically determined.}.

\subsubsection{OPIEC} 
OPIEC \cite{gashteovski2019opiec} is a large-scale dataset generated using an open information extraction (OIE) system applied to the text of Wikipedia. For each sentence in Wikipedia, OPIEC extracts one or more \emph{(subject, negation, relation, object, time, space)} tuples. The tuples are dense, structured versions of the original Wikipedia sentences. 

We retrieve OPIEC tuples by performing soft sentence matching between the knowledge sentences in WoW and OPIEC's source sentences. There are some mismatches between sentences in these two datasets because the Wikipedia dumps used are not exactly the same, due to the continuous editing of Wikipedia contributors. We use Sentence-BERT \footnote{all-MiniLM-L6-v2 checkpoint from \url{https://www.sbert.net/}.} \cite{reimers-gurevych-2019-sentence} to encode sentences from WoW and OPIEC that appear in the same Wikipedia article and consider sentence pairs with cosine similarity larger than $0.9$ as a match.

\subsubsection{Semantic frames} 
Semantic frames (sem. frm. or s.f.) capture the core semantics of sentences, such as who did what, when, and where. Previous work on OIE has studied the use of semantic role labeling (SRL)-based knowledge tuples \cite{christensen2011analysis}. Therefore, we use SRL results as complementary structural knowledge to the OPIEC tuples. We use the semantic roles parsed by spaCy's\footnote{\url{https://spacy.io/}} SRL pipeline for each WoW knowledge sentence and use templates to map the results into \emph{(subject, relation, object, time, space)} tuples.

\begin{table*}[t]
\setlength{\tabcolsep}{4.8pt} 
\renewcommand{\arraystretch}{1.0} 
\begin{center}
\small
    \begin{tabular}{  l  l  l  l  l  l}
    \hline
    \textbf{Multi-Source WoW} & Train & Valid Seen & Valid Unseen & Test Seen & Test Unseen \\ \hline
    Number of utterances & 166787 & 8909 & 8806 & 8715 & 8782\\ 
    Number of dialogues  & 18430 & 981 & 967 & 965 & 968\\ 
    Number of topics  & 1247 & 545 & 54 & 533 & 58\\ 
    Avg turns per dialogue  & 9.05 & 9.08 & 9.13  & 9.03 & 9.07\\ 
    \% of Wizard turns with knowledge & 61.8 & 62.5 & 65.0 & 62.8 & 61.8\\ 
    Avg non-zero \# of knowledge per utterance  & 5.0 & 5.0 & 5.1 & 5.0 & 4.9 \\ 
    Avg \# of (gold) Knowledge per Utterance & 13.8 (1.55) & 13.4 (1.49) & 14.1 (1.59) & 13.9 (1.55) & 15.2 (1.59) \\
    \% of (gold) OPIEC & 64.9 (57.4) & 65.2 (59.2) & 57.9 (51.5) & 65.4 (58.2) & 67.4 (58.7)\\
    \% of (gold) semantic frame & 4.55 (9.87) & 4.17 (8.73) & 3.82 (8.85) & 4.37 (8.98) & 5.08 (12.1) \\
    \% of (gold) Wikidata & 17.2 (13.9) & 17.2 (12.9) & 21.7 (16.9) & 17.2 (13.9) & 16.8 (12.4)\\
    \% of (gold) Wikipedia & 13.4 (18.8) & 13.5 (19.2) & 16.6 (22.8) & 13.0 (18.9) & 10.7 (16.7)\\
    
    \hline
    \end{tabular}
    \vspace{-0.5em}
    \caption{Statistics of our Multi-Source Wizard of Wikipedia.} \label{tab:statistics}
    \vspace{-1.5em}
\end{center}
\end{table*}

\subsubsection{Wikidata} 
Wikidata (w.d.) is a large-scale knowledge base containing triplets grounded in Wikipedia articles. It contains \emph{(subject, relation, object)} triplets that relate one Wikipedia concept to another. We retrieve triplets from Wikidata using the following steps:

\paragraph{Entity and noun phrase detection.}
We use spaCy to extract entities and noun phrases from each sentence, filtered with NLTK\footnote{\url{https://www.nltk.org/}} stop words.

\paragraph{Entity linking.}
We pass each extracted entity or noun phrase, along with its original sentence as context, to a dense retrieval-based entity linker \cite{wu-etal-2020-scalable} to obtain the corresponding Wikidata entities. We take the top-1 Wikidata entity candidate for each extracted entity or noun phrase.

\paragraph{Triplet retrieval.}
We retrieve all Wikidata triplets that contain at least one of the linked entities.

\paragraph{Triplet filtering.}
We keep only triplets whose subjects and objects both match the corresponding WoW knowledge sentence, requiring one of the following conditions: 
\begin{itemize}
  \item both the subject and object entities in the triplet appear in the set of linked Wikidata entities extracted from the WoW sentence, or
  \item the subject and object both match the WoW sentence using a fuzzy matcher\footnote{\url{https://spacy.io/universe/project/spaczz}} with score higher than $0.95$.
\end{itemize}

\paragraph{Coverage-based filtering.}
For each WoW knowledge sentence, we keep only triplets whose entities cover more than 75\% of the extracted entity set of the sentence.

\subsection{Post-processing} \label{sec:post_processing}
We perform post-processing to ensure the quality of the retrieved tuples and create complementary and \textit{disjoint partitions} of knowledge sources by filtering out semantically incomplete and redundant tuples.

\subsubsection{Filtering of Retrieved Tuples}
\label{sec:joint_tuple_filtering}
\paragraph{Ensuring sufficient semantic coverage.} In order to minimize information loss when we replace each WoW sentence with the retrieved tuples, we filter out those sentences whose retrieved tuples can only partially cover their semantics. We tokenize and lemmatize each WoW knowledge sentence and remove punctuation and stop words to create a bag of words $U$ for each sentence. Then we concatenate all retrieved tuple elements from the three sources to create a pseudo-sentence and its corresponding bag of words $S$. We only keep those sets of tuples whose $S$ covers more than 60\% of $U$.

This filtering step results in the creation of a fourth, supplementary knowledge source: Wikipedia sentences (w.p.) from the original WoW that could not be adequately captured by tuples from the other three sources.

\paragraph{Deduplication of redundant tuples.}  Since the knowledge tuples are retrieved independently from the three sources, some are redundant with each other. Since our goal is to partition the knowledge into complementary sources, we perform deduplication to remove redundant tuples. Given the full-coverage bag of words $A = U \cap S$, we want to select the minimum number of tuples that maximally cover $A$. We formulate this goal as a set-cover problem, which is NP-Complete, and apply its 2-approximation algorithm to select the minimum set of tuples that covers the semantics of the original WoW sentence to the same extent as the full set of retrieved tuples.

\subsubsection{Grounding Gold Knowledge Tuples to Utterances}
In the original WoW dataset \cite{dinan2019wizard}, each Wizard utterance has at most one gold support knowledge sentence. However, in our Ms.WoW dataset, each original WoW support knowledge sentence is decomposed into multiple knowledge tuples (see Table \ref{tab:example} for an example). Since the semantics of the sentence is spread among these tuples, some tuples derived from the WoW knowledge sentence may contain extra information not found in the corresponding Wizard utterance. Therefore, we use the same set-cover approach as in Section \ref{sec:joint_tuple_filtering} to select those knowledge tuples that are grounded by their utterances; we refer to these grounded tuples as ``gold" tuples. 

\subsection{Dataset Statistics} \label{sec:statistics}
Table \ref{tab:statistics} shows the statistics of our Ms.WoW dataset. OPIEC tuples are parsed from the entire Wikipedia, so they have the most coverage. In contrast, semantic frame tuples are much fewer in number due to our template-based matching rule. 

\begin{table}[t]
\setlength{\tabcolsep}{4pt} 
\renewcommand{\arraystretch}{1.0} 
\begin{center}
\small
    \begin{tabular}{  l | l l l l l l l } \hline
     & sbj & neg & rel & obj & tmp & spa & total\\ \hline
    OPIEC  & 2.2 & 1.0 & 3.3 & 2.8 & 2.1 & 2.3 & 8.7\\ 
    Sem. frm.  & 6.3 & $-$ & 1.0 & 14.3 & 3.9 & 8.2 & 22.1\\ 
    Wikidata & 1.7 & $-$ & 2.2 & 1.7 & $-$ & $-$ & 5.6\\ 
    Wikipedia  & $-$ & $-$ & $-$ & $-$ & $-$ & $-$ & 24.9\\
    \hline
    \end{tabular}
    \vspace{-1.0em}
    \caption{Average of number of words per non-empty knowledge attribute.} \label{tab:knowledge_component_lens}
    \vspace{-1.0em}
\end{center}
\end{table}

Despite originating from the same Wikipedia sentences, tuples derived from different sources have significantly different length attributes (see Table \ref{tab:example} for examples). As Table \ref{tab:knowledge_component_lens} shows, different knowledge sources yield knowledge with different numbers of components: while Wikidata tuples only have \emph{(subject, relation, object)}, OPIEC tuples also have \emph{negation, time}, and \emph{space}. The lengths of each knowledge type's attributes are also significantly different: semantic frame tuples have single-word relations with long subjects and objects, while OPIEC tuples have longer relations with shorter subjects and objects. In addition, while semantic frames retain the original sentence tokens, OPIEC and Wikidata decompose the original sentences into multiple pieces.

\subsection{Quality assurance via dialogue response generation} \label{sec:quality_assurance}
Since Ms.WoW is a new dataset derived from WoW but designed for a different purpose, we want to ensure that the knowledge tuples in Ms.WoW sufficiently retain the information in the WoW knowledge sentences. Table \ref{tab:response_generation_results_zero_shot} shows that response generators using our full Ms.WoW. (all sources) perform comparably to those using the original WoW dataset: our ROUGE scores and utterance F1 are comparable, and our unigram multi-source knowledge F1 is close to that of the original WoW setting (metric descriptions are found in Section \ref{sec:experimental_design}). This confirms that our Ms.WoW covers the knowledge needed to support a dialogue model to generate high-quality responses with limited information loss compared to WoW.

\section{Dialogue Knowledge Plug-and-Play} \label{sec:experiments}
Having collected Ms.WoW, we apply it as a test bed to study \textit{dialogue knowledge plug-and-play} via multi-source dialogue knowledge selection and response generation. We employ a baseline approach to demonstrate the basic characteristics and challenges of \textit{dialogue knowledge plug-and-play}.

\begin{table}[t]
\begin{center}
\small
    \begin{tabular}{  l | l l l}
    \hline
     &  \multicolumn{3}{l}{\textit{Test Seen + Unseen}} \\
    \textbf{Training} & \textbf{P} & \textbf{R} & \textbf{F1} \\ \hline
    Full Kn. & 0.384 & 0.380 & 0.382 \\ \hline
    $-$ OPIEC & 0.368 & 0.274 & 0.314 \\ 
    $-$ Sem. frm. & 0.404 & 0.333 & 0.365 \\ 
    $-$ Wikidata & 0.446 & 0.303 & 0.361  \\ 
    $-$ Wikipedia & 0.497 & 0.319 & 0.389  \\ \hline

    \hline
    \end{tabular}
    \vspace{-0.5em}
    \caption{Dialogue knowledge selection performance on Ms.WoW test set (seen + unseen).} \label{tab:knowlege_selection_results}
    \vspace{-1.5em}
\end{center}
\end{table}

\begin{table*}[t]
\setlength{\tabcolsep}{4pt} 
\renewcommand{\arraystretch}{1.0} 
\begin{center}
\small
    \begin{tabular}{  l | l l l | l l l | l l l | l l l }
    \hline
     &  \multicolumn{3}{l}{\textit{OPIEC}} & \multicolumn{3}{l}{\textit{Sem. frm.}} & \multicolumn{3}{l}{\textit{Wikidata}} & \multicolumn{3}{l}{\textit{Wikipedia}} \\
    \textbf{Training} & \textbf{P} & \textbf{R} & \textbf{F1} & \textbf{P} & \textbf{R} & \textbf{F1} & \textbf{P} & \textbf{R} & \textbf{F1} & \textbf{P} & \textbf{R} & \textbf{F1}\\ \hline
    Full Knowledge & 0.340 & 0.347 & 0.343 & 0.591 & 0.639 & 0.614  & 0.301 & 0.397 & 0.342 & 0.550 & 0.321 & 0.406 \\ \hline
    $-$ OPIEC & \emph{0.301} & \emph{0.194} & \emph{0.236}  & 0.549 & 0.647 & 0.594 & 0.256 & 0.180 & 0.211 & 0.459 & 0.384 & 0.418 \\ 
    $-$ Sem. frm. & 0.352 & 0.275 & 0.309 & \emph{0.580} & \emph{0.585} & \emph{0.583} & 0.360 & 0.270 & 0.309 & 0.460 & 0.418 & 0.438 \\ 
    $-$ Wikidata & 0.401 & 0.265 & 0.319 & 0.587 & 0.640 & 0.612 & \emph{0.277} & \emph{0.087} & \emph{0.133} & 0.505 & 0.391 & 0.441 \\ 
    $-$ Wikipedia & 0.473 & 0.244 & 0.322 & 0.569 & 0.701 & 0.628 & 0.436 & 0.268 & 0.332 & \emph{0.519} & \emph{0.376} & \emph{0.436 } \\
    \hline
    \end{tabular}
    \vspace{-0.5em}
    \caption{Dialogue knowledge selection performance on Ms.WoW test set (seen + unseen) by knowledge source. All knowledge sources are present during testing, simulating the scenario where a new knowledge source becomes available at test time.} \label{tab:type_wise_knowlege_selection_results_mini}
    \vspace{-1.5em}
\end{center}
\end{table*}

\begin{table*}[t]
\setlength{\tabcolsep}{4pt} 
\renewcommand{\arraystretch}{1.0} 
\begin{center}
\small
    \begin{tabular}{  l | l l l | l l l | l l l | l l l }
    \hline
     &  \multicolumn{3}{l}{\textit{OPIEC}} & \multicolumn{3}{l}{\textit{Sem. frm.}} & \multicolumn{3}{l}{\textit{Wikidata}} & \multicolumn{3}{l}{\textit{Wikipedia}} \\
    \textbf{Training \& Testing} & \textbf{P} & \textbf{R} & \textbf{F1} & \textbf{P} & \textbf{R} & \textbf{F1} & \textbf{P} & \textbf{R} & \textbf{F1} & \textbf{P} & \textbf{R} & \textbf{F1}\\ \hline
    $-$ OPIEC & $-$ & $-$ & $-$ & 0.584 & 0.584 & 0.584 & 0.293 & 0.199 & 0.237 & 0.459 & 0.395 & 0.425 \\ 
    $-$ Sem. frm. & 0.349 & 0.274 & 0.307 & $-$ & $-$ & $-$ & 0.352 & 0.266 & 0.303 & 0.452 & 0.413 & 0.431 \\ 
    $-$ Wikidata & 0.389 & 0.268 & 0.317 & 0.579 & 0.623 & 0.600 & $-$ & $-$ & $-$ & 0.431 & 0.322 & 0.368 \\ 
    $-$ Wikipedia & 0.472 & 0.226 & 0.306 & 0.575 & 0.687 & 0.626 & 0.452 & 0.245 & 0.318 & $-$ & $-$ & $-$ \\ 
    \hline
    \end{tabular}
    \vspace{-0.5em}
    \caption{Dialogue knowledge selection performance on the Ms.WoW test set (seen + unseen), excluding the ablated knowledge source for each model; both training and testing are conducted with one knowledge source missing, simulating the scenario where one knowledge source never becomes available.} \label{tab:ablated_knowledge_selection_mini}
    \vspace{-1.5em}
\end{center}
\end{table*}

\subsection{Experimental Design} \label{sec:experimental_design}

Retraining a dialogue model each time a new knowledge source becomes available is computationally costly and requires extra engineering effort. \textit{Dialogue knowledge plug-and-play} examines the ability of a pretrained dialogue model to adapt to support knowledge from new sources in a zero-shot fashion.

We use Ms.WoW to simulate the realistic scenario where an additional knowledge source becomes available after the dialogue model has already been trained. We test a model's adaptability to new knowledge sources by ablating one of the Ms.WoW sources from the available candidate knowledge during training and then test the ablated model with the full-knowledge test set; the missing knowledge source becomes available only at test time (Tables \ref{tab:type_wise_knowlege_selection_results_mini} \& \ref{tab:response_generation_results_zero_shot}). The goal of the \textit{dialogue knowledge plug-and-play} challenge is to reduce the difference between the test performance of each knowledge-ablated model and the test performance of a model trained on the full-knowledge dataset; the challenge prefers models that can quickly adapt to make use of the previously unseen knowledge source. 

For response generation, we compare the knowledge-ablated models to two upper bounds. First, we train a model on the full set of available knowledge tuples\footnote{Note that ``full knowledge" refers to the full set of candidate knowledge tuples available for each turn, derived from the corresponding full set of WoW knowledge sentences, which have already been filtered from a large pool of millions of Wikipedia articles using an information retrieval module described by \citet{dinan2019wizard}} (\textit{Ms.WoW full knowledge}), simulating a model that is retrained when the new knowledge source becomes available. Second, we experiment with using gold (i.e. utterance-grounded) knowledge tuples only, simulating the scenario where a ``perfect" knowledge selector is available (\textit{Ms.WoW gold knowledge}) and providing an upper bound on the effect of knowledge selection performance on response generation.

\begin{table}[t]
\small
\begin{center}
\setlength{\tabcolsep}{5pt} 
    \begin{tabular}{p{0.9\linewidth} }
    \hline
    \textbf{Prompt} \\ \hline
    The following is the conversation between the ``Wizard", a knowledgeable speaker who can access Wikipedia knowledge sentences to chat to with the ``Apprentice", who does not have access to Wikipedia.
    The conversation topic is \{\{topic\}\} and the persona setting of the Wizard is ``\{\{persona\}\}".\\ \hdashline
    This is their conversation history: \\
    \{\{speaker 1\}\}: \{\{utterance 1.1\}\} \\
    \{\{speaker 2\}\}: \{\{utterance 2.1\}\} \\
    \{\{speaker 1\}\}: \{\{utterance 1.2\}\} \\
    ... \\ \hdashline
    Here is some retrieved Wikipedia knowledge for the Wizard. \\
    Some of the knowledge is in the tuple form, such as (subject, negation, relation, object, time, space) or (subject, relation, object). \\
    The Wizard can choose any subset of the following knowledge. It's also allowed to not choose any of them. \\ 
    \{\{(subject 1, relation 1, object 1)\}\} \\
    ... \\
    \hdashline
    Given the knowledge above, make a very brief, such as one sentence, natural response for the Wizard. \\
    Not all information in the chosen knowledge has to be used in the response. \\
    The Wizard's response is:\\ \hline
    \end{tabular}
    \vspace{-0.5em}
    \caption{Prompt for Vicuna-13B dialogue response generation using Ms.WoW full knowledge.} \label{tab:prompt}
    \vspace{-1.5em}
\end{center}
\end{table}

\begin{table}[t]
\begin{center}
\small
    \begin{tabular}{  l  l  l }
    \hline
    Hyper-parameter & Selector & Generator \\ \hline
    Learning rate & 1e-5 & 5e-5 \\
    Batch size & 16 & 12 \\
    Epochs & 10 & 10 \\
    Max sequence length & 512 & 512 \\
    \hline
    \end{tabular}
    \vspace{-0.5em}
    \caption{Fine-tuned hyper-parameters.} \label{tab:hyper_parameters}
    \vspace{-1.5em}
\end{center}
\end{table}

\subsection{Baseline Approaches}
\label{sec:baseline_approach}
\subsubsection{Fine-tuned Models}
\paragraph{Input encoding.} 
We create the input sequences by concatenating up to the last five utterances in the conversation history ($u_i$), speaker roles ($s_i$), and the support knowledge sentences and tuples ($k_j$) for each Wizard dialogue turn. Each support knowledge subsequence is prepended with a special token <kg>. We feed the input sequence to a pre-trained language model. The input sequence $x$ can be written as:
\begin{equation}
\label{eq1}
x = [s_1: u_1, ..., s_5: u_5, \text{<kg>}, k_1, ..., \text{<kg>}, k_j]
\end{equation}

\paragraph{Knowledge selector.} 
We use a Roberta-base \cite{liu2019roberta} model ($Roberta$), followed by a 2-layer feed-forward network ($MLP$). We take each support knowledge tuple's corresponding <kg> token representation $H^{<kg>}_j$ as its knowledge representation for knowledge selection.  Utterances without any candidate knowledge are skipped.
\begin{equation}
\begin{aligned}
H &= Roberta(x) \\
y_j &= softmax(MLP(H^{<kg>}_j))
\end{aligned}
\end{equation}

\paragraph{Response generator.}
We use a BART-base \cite{lewis2019bart} model ($BART$) to perform a standard response generation using the same input $x$ from Equation \ref{eq1}: $u' = BART(x)$.

\subsubsection{Large Language Model}
We also prompt an LLM in a zero-shot fashion for the response generation task. We use Vicuna-13B \cite{vicuna2023}, a 13-billion-parameter LLaMA \cite{touvron2023llama} model fined-tuned on 70k user-shared conversation samples. 
Table \ref{tab:prompt} shows the prompt we use.

\begin{table*}[t]
\setlength{\tabcolsep}{4pt} 
\renewcommand{\arraystretch}{1.0} 
\begin{center}
\small
    \begin{tabular}{  l | l | l l l | l | l l l }
    \hline
    \textbf{Configurations} & \textbf{Training} & R-1 & R-2 & R-L & F1 & K-P & K-R & K-F1 \\ \hline
    No knowledge & No knowledge & 0.189 & 0.043 & 0.159 & 0.207  & $-$ & $-$ & $-$ \\ \hline \hline
    WoW Full knowledge & WoW Full knowledge & 0.261 & 0.101 & 0.225 & 0.265 & 0.502 & 0.162 & 0.245 \\ \hline
    Ms.WoW Full knowledge & Ms.WoW Full knowledge & 0.259 & 0.094 & 0.222 & 0.264 & 0.460 & 0.159 & 0.236\\ 
    & $-$ OPIEC & 0.247 & 0.084 & 0.212 & 0.251 & 0.433 & 0.146 & 0.219 \\
    & $-$ sem. frm. & 0.256 & 0.093 & 0.219 & 0.260 & 0.448 & 0.158 & 0.234 \\
    & $-$ Wikidata & 0.256 & 0.093 & 0.220 & 0.261 & 0.440 & 0.154 & 0.228 \\
    & $-$ Wikipedia & 0.251 & 0.089 & 0.215 & 0.256 & 0.459 & 0.164 & 0.242 \\
    \hline \hline
    WoW Gold knowledge & WoW Gold knowledge & 0.317 & 0.150 & 0.278 & 0.317 & 0.387 & 0.528 & 0.446 \\ \hline
    Ms.WoW Gold knowledge & Ms.WoW Gold knowledge & 0.322 & 0.149 & 0.280 & 0.321 & 0.311 & 0.576 & 0.404\\ 
    &$-$ OPIEC & 0.306 & 0.133 & 0.265 & 0.302 & 0.302 & 0.560 & 0.392\\
    &$-$ sem. frm. & 0.321 & 0.147 & 0.280 & 0.321 & 0.316 & 0.582 & 0.409\\
    &$-$ Wikidata & 0.320 & 0.145 & 0.279 & 0.319 & 0.313 & 0.580 & 0.406 \\
    &$-$ Wikipedia & 0.319 & 0.145 & 0.277 & 0.318 & 0.311 & 0.585 & 0.406 \\
    \hline
    \end{tabular}
    \vspace{-0.5em}
    \caption{Response generation performance on test set  (seen + unseen). All knowledge sources are present during testing, simulating the scenario where a new knowledge source becomes available at test time. Full knowledge refers to no knowledge selection, where all available candidate knowledge is used; gold knowledge refers to oracle knowledge selection.} \label{tab:response_generation_results_zero_shot}
    \vspace{-1.5em}
\end{center}
\end{table*}

\begin{table*}[t]
\setlength{\tabcolsep}{4pt} 
\renewcommand{\arraystretch}{1.0} 
\begin{center}
\small
    \begin{tabular}{  l | l | l l l | l | l l l}
    \hline
    \textbf{Configurations} & \textbf{Training \& Testing} & R-1 & R-2 & R-L & F1 & K-P & K-R & K-F1\\ \hline
    Ms.WoW Full knowledge & $-$ OPIEC & 0.238 & 0.077 & 0.203 & 0.245 & 0.281 & 0.098 & 0.145\\
    &$-$ sem. frm. & 0.251 & 0.088 & 0.216 & 0.256 & 0.428 & 0.150 & 0.222 \\
    &$-$ Wikidata  & 0.257 & 0.093 & 0.221 & 0.261 & 0.331 & 0.116 & 0.171 \\
    &$-$ Wikipedia & 0.245 & 0.083 & 0.209 & 0.250 & 0.361 & 0.126 & 0.186 \\
    \hline \hline
    Ms.WoW Gold knowledge & $-$ OPIEC & 0.253 & 0.094 & 0.217 & 0.260 & 0.198 & 0.378 & 0.260\\
    &$-$ sem. frm. & 0.306 & 0.134 & 0.266 & 0.309  & 0.274 & 0.500 & 0.354\\
    &$-$ Wikidata & 0.317 & 0.144 & 0.277 & 0.316 & 0.305 & 0.562 & 0.396\\
    &$-$ Wikipedia & 0.290 & 0.119 & 0.250 & 0.293 & 0.237 & 0.440 & 0.308\\
    \hline
    \end{tabular}
    \vspace{-0.5em}
    \caption{Response generation performance on the knowledge-ablated test set (seen + unseen). Each model is trained and tested on the same knowledge sources (full or gold), simulating the scenario where one knowledge source never becomes available.} \label{tab:response_generation_results_ablation}
    \vspace{-1.5em}
\end{center}
\end{table*}

\begin{table*}[t]
\setlength{\tabcolsep}{4pt} 
\renewcommand{\arraystretch}{1.0} 
\begin{center}
\small
    \begin{tabular}{  l | l | l l l | l | l l l }
    \hline
    \textbf{Configurations} & \textbf{Testing} & R-1 & R-2 & R-L & F1 & K-P & K-R & K-F1 \\ \hline
    No knowledge & No knowledge & 0.187 & 0.031 & 0.147 & 0.202 & $-$ & $-$ & $-$ \\ \hline \hline
    WoW Full knowledge & WoW Full knowledge & 0.202 & 0.046 & 0.153 & 0.216 & 0.216 & 0.184 & 0.199 \\ \hline
    Ms.WoW Full knowledge & Ms.WoW Full knowledge & 0.196 & 0.044 & 0.149 & 0.212 & 0.382 & 0.260 & 0.310 \\ 
    & $-$ OPIEC & 0.185 & 0.034 & 0.141 & 0.199 & 0.258 & 0.157 & 0.195 \\
    & $-$ sem. frm. & 0.189 & 0.038 & 0.143 & 0.203 & 0.343 & 0.225 & 0.272 \\
    & $-$ Wikidata & 0.196 & 0.043 & 0.149 & 0.211 & 0.379 & 0.256 & 0.306 \\
    & $-$ Wikipedia  & 0.196 & 0.043 & 0.149 & 0.210 & 0.373 & 0.247 & 0.297 \\
    \hline \hline
    WoW Gold knowledge & WoW Gold knowledge & 0.218 & 0.053 & 0.170 & 0.226 & 0.049 & 0.086 & 0.062 \\ \hline
    Ms.WoW Gold knowledge & Ms.WoW Gold knowledge & 0.230 & 0.061 & 0.176 & 0.236 & 0.114 & 0.351 & 0.172 \\ 
    &$-$ OPIEC  & 0.193 & 0.038 & 0.147 & 0.205 & 0.072 & 0.219 & 0.109 \\
    &$-$ sem. frm. & 0.215 & 0.051 & 0.165 & 0.223 & 0.095 & 0.289 & 0.144 \\
    &$-$ Wikidata & 0.220 & 0.056 & 0.170 & 0.226 & 0.108 & 0.325 & 0.162 \\
    &$-$ Wikipedia & 0.224 & 0.057 & 0.172 & 0.230 & 0.110 & 0.332 & 0.165 \\
    \hline
    \end{tabular}
    \vspace{-0.5em}
    \caption{Vicuna-13B response generation performance on the test set (seen + unseen).} \label{tab:response_generation_results_llm}
    \vspace{-1.5em}
\end{center}
\end{table*}

\subsection{Experimental Details}
\paragraph{Fine-tuned Models.} We use the Roberta-base (125M parameters) and BART-base (139M parameters) models from Huggingface\footnote{\url{https://huggingface.co/models}}. We mostly use the default hyper-parameters (see Table \ref{tab:hyper_parameters}). We use a single Nvidia Tesla V100S GPU for model training and testing. Each dialogue knowledge selector and dialogue generator takes approximately 3 hours for training and a few minutes for inference.

\paragraph{Large Language Model.} We use the Vicuna-13B-v1.1 LLM from Huggingface. We set the generation temperature to 0.7. It takes approximately 30 hours to perform inference on the test set (seen + unseen) for each configuration using 4 Nvidia Tesla V100S GPUs.

\section{Experimental Results and Analysis}
\label{sec:zero_shot_experiments}

Tables \ref{tab:knowlege_selection_results}, \ref{tab:type_wise_knowlege_selection_results_mini}, and \ref{tab:ablated_knowledge_selection_mini} show the performance of our Roberta-based knowledge selector; Tables \ref{tab:response_generation_results_zero_shot} and \ref{tab:response_generation_results_ablation} show the performance of our BART-based response generator; and Table \ref{tab:response_generation_results_llm} shows the performance of the LLM response generator.

To measure the dialogue response generation performance, in addition to ROUGE scores \cite{lin-2004-rouge} compared to the gold response, we follow \citet{dinan2019wizard} in reporting the unigram F1 of the generated response with the gold response, as well as unigram precision, recall and F1 (K-P, K-R \& K-F1) of the generated response with all available (i.e. non-ablated) candidate knowledge.

\subsection{Full-knowledge vs. Zero-shot Adaptation} Unsurprisingly, there is a significant difference between the full-knowledge model (i.e. retrained with each new knowledge source) and the zero-shot adapted models that have one knowledge source ablated during training (Tables \ref{tab:knowlege_selection_results} and \ref{tab:response_generation_results_zero_shot}). This performance gap generally increases as the ablated knowledge source occupies a larger proportion of the overall available knowledge. 

This trend is clearer when we separately examine the knowledge selectors' performances on each knowledge source in Table \ref{tab:type_wise_knowlege_selection_results_mini} (diagonal entries vs. full knowledge). In general, an ablated model's recall score on the newly available knowledge source is dramatically lower than that of the full-knowledge model; the distribution of the new knowledge source is not recognized as usable support knowledge. This observation clearly shows the significant performance gap between the full-knowledge model and the zero-shot adapted models, which is exactly the gap that our \textit{dialogue knowledge plug-and-play} challenge aims to highlight as a target for improvement.

Interestingly, Table \ref{tab:type_wise_knowlege_selection_results_mini} (non-diagonal entries vs. full knowledge) also shows that a new knowledge source becoming available improves the model's knowledge selection performance on some of the previously available knowledge sources, demonstrating that there is some synergy among knowledge from different sources. 

The only exception to the observations above is the supplementary Wikipedia sentence source, consisting of WoW sentences that could not be adequately covered by our three other sources. We suspect this is because the pre-trained language model we use, Roberta \cite{liu2019roberta}, is already extensively trained on Wikipedia articles, which makes zero-shot adaptation back to the Wikipedia sentences trivial. 

\subsection{Differences among Knowledge Sources} 

There is a clear difference in difficulty among knowledge sources. Since semantic frame tuples are extracted using high-precision, human-engineered templates, all models perform significantly better on semantic frame tuples than other knowledge types (Table \ref{tab:type_wise_knowlege_selection_results_mini}). 

Different knowledge sources also have different usefulness in response generation. As Table \ref{tab:response_generation_results_zero_shot} and \ref{tab:response_generation_results_llm} show, Wikidata knowledge triplets are not as helpful as the other knowledge sources. This may be because Wikidata triplets are generally short (Section \ref{sec:statistics}), making them less informative than the other knowledge sources.

\subsection{``More is Better'' in Zero-shot Settings} 

Tables \ref{tab:ablated_knowledge_selection_mini} and \ref{tab:response_generation_results_ablation} show the knowledge selection and response generation performance of our models when one knowledge source is ablated from both training and testing, simulating a scenario where that knowledge source never becomes available. Comparing these results with Tables \ref{tab:type_wise_knowlege_selection_results_mini} and \ref{tab:response_generation_results_zero_shot}, respectively, we see that introducing additional knowledge sources, even in a zero-shot fashion, mostly benefits, rather than hurts, knowledge selection and response generation, supporting the claim that ``more (knowledge sources) is better'' \cite{wu2021more}. This is a promising result for \textit{dialogue knowledge plug-and-play}, which challenges models to be robust to new knowledge sources. This phenomenon is also relevant to the in-context-learning \cite{brown2020language} of LLMs, where LLMs learn from new inputs in a few-shot manner.



\subsection{LLMs for Dialogue Knowledge Plug-and-Play} 
LLM-based response generation task can be considered an extreme scenario of zero-shot prediction, where no in-domain training is conducted at all. Compared to the fine-tuned models, which have the opportunity to learn the speaking style and statistical distribution of the conversationalists, the Vicuna-generated zero-shot responses are significantly longer: mean token length of 42.5 vs. 21.3 from the BART-based model on the Ms.WoW full knowledge test set; the mean target response length is 24.5 tokens. The Vicuna-generated outputs have less overlap with the corresponding gold responses, but a higher unigram overlap with the input knowledge in the full-knowledge setting (Table \ref{tab:response_generation_results_llm}), indicating that the LLM is able to generate utterances relevant to the input knowledge when sufficient knowledge is given. Surprisingly, when provided with gold knowledge only, the LLM seems not to use the provided knowledge to the same extent, and we see a decrease in performance, in contrast with the BART-based model that improved with gold knowledge. However, similar to the BART-based model, we still observe that more knowledge sources provided at test time significantly improves the LLM's response generation performance across all metrics.

\section{Conclusion}
We introduce the Ms.WoW dataset of multi-source support knowledge for open-domain dialogue generation, with knowledge tuples partitioned into disjoint sources and grounded at the utterance level. We further introduce the \textit{dialogue knowledge plug-and-play} challenge, where a trained dialogue system must adapt to a new knowledge source at test time. Our baseline experiments demonstrate how future works can use Ms.WoW to study how dialogue models generalize to new knowledge sources.

\section*{Limitations}
The source of our Ms.WoW knowledge tuples are from OPIEC \cite{gashteovski2019opiec}, semantic frames and Wikidata, all of whose tuples are automatically collected using trained models or rules without direct human annotation. Therefore these knowledge tuples inevitably inherit the noise from their knowledge sources, despite we show they sufficiently retain the semantics of the original Wikipedia knowledge sentences (Section \ref{sec:quality_assurance}).

\section*{Ethics Statement}
As we previously explained, the source of our Ms.WoW knowledge tuples are from OPIEC \cite{gashteovski2019opiec}, semantic frames and Wikidata, all of whose tuples are automatically collected using trained models or rules without direct human annotation. All these external artifacts are properly cited and processed according to their licenses and requirements. In other words, we do not expect to introduce any additional sensitive issues to our work.
\section{Bibliographical References}\label{sec:reference}

\bibliographystyle{lrec-coling2024-natbib}
\bibliography{custom}

\bibliographystylelanguageresource{lrec-coling2024-natbib}
\bibliographylanguageresource{languageresource}

\end{document}